\documentclass[runningheads]{llncs}
\usepackage[T1]{fontenc}
\usepackage{graphicx,subcaption}
\usepackage{booktabs}
\begin{document}
\title{The Impact of Featuring Comments in Online Discussions}
%
%
\author{Cedric Waterschoot\inst{1}\orcidID{0000-0003-4903-2604} \and
Ernst van den Hemel\inst{1}\and
Antal van den Bosch\inst{2}}
\authorrunning{C. Waterschoot et al.}
%
\institute{KNAW Meertens Instituut, Amsterdam, The Netherlands \\
\email{\{cedric.waterschoot, ernst.van.den.hemel2\}@meertens.knaw.nl}\\
 \and
Institute for Language Sciences, Utrecht University, Utrecht, The Netherlands\\
\email{a.p.j.vandenbosch@uu.nl}}
\maketitle              
\begin{abstract}
A widespread moderation strategy by online news platforms is to feature what the platform deems high quality comments, usually called editor picks or featured comments. In this paper, we compare online discussions of news articles in which certain comments are featured, versus discussions in which no comments are featured. We measure the impact of featuring comments on the discussion, by estimating and comparing the quality of discussions from the perspective of the user base and the platform itself. Our analysis shows that the impact on discussion quality is limited. However, we do observe an increase in discussion activity after the first comments are featured by moderators, suggesting that the moderation strategy might be used to increase user engagement and to postpone the natural decline in user activity over time.

\end{abstract}
\section{Introduction}
Commenting on online news articles has been growing in popularity. However, many potential participants refrain from commenting due to the negativity associated with only discussions \cite{vanderlinden2024}. How to foster constructive debate in the comment section? This question is of increasing concern for online media outlets. For at least two decades, online news platforms have been struggling to curtail dark participation and trolling in comment spaces \cite{Quandt2018}. In recent years, however, the moderator received an increasingly wider range of tasks besides merely deleting undesired user content \cite{Waterschoot2024}. The moderator is now also tasked with recognizing and promoting \textit{good} comments. Featuring quality comments as a norm-setting strategy has become widespread among large online outlets such as, for example, the New York Times and the Guardian \cite{Wang2022,Diakopoulos2015b}. Presented as examples of constructive discussion among users, certain comments are pinned to a highly visible position within the comment interface. However, much remains unclear regarding the actual impact and implications of this moderation strategy on the discussion. What happens in the comment space when moderators start promoting certain comments and commenters? 

In this paper, we analyze discussions in which moderators performed the moderation strategy of featuring high quality comments, by comparing them to discussions in which no comments were featured. By further splitting online discussions, either with or without featured comments, in before and after subgroups based on the featuring time of comments in the data, we are able to pinpoint differences in discussion quality and activity between the control and the featured content discussions. 
Overall, discussion quality was unaffected. Quality as estimated from the editorial perspective was not impacted by the presence of featured comments. Due to the fact that the control discussions' activity dwindled down faster over time, we end our analysis by hypothesizing whether featuring comments can be used to extend activity on the discussion platform. Indeed, we observe more engagement in terms of posts and involved users in discussions with featured comments.

\section{Background}
The task of content moderation is defined as the screening of user-generated content to assess the appropriateness for the given platform \cite{Roberts2017,Gillespie2018}. The practice has been evolving to address the needs of a growing, contemporary online community. This development expanded the task set of the moderator, who needs to swiftly make interpretative moderation choices \cite{Paasch2022}. 
Practically speaking, the moderation task has been described as a gatekeeping role \cite{Wolfgang2018}. This function is twofold. First, moderators ought to keep the comment space clean of unwanted content \cite{Paasch2022}. Described under the umbrella of dark participation, this content can take the form of trolling, cyberbullying or even organized misinformation campaigns \cite{Quandt2018,Lewandowsky2017,VanderLinden2017}. The moderator is charged with deleting or reducing the visibility of such content \cite{Gillespie2022}. Additionally, coping with these negative influences required the platforms to expand the online content moderation practice \cite{Wintterlin2020}. This expansion, among other things, led to the promotion of \textit{good} content \cite{Wolfgang2018,Diakopoulos2015b}. The bulk of the literature on content moderation focuses on the bad and unwanted comments and actors within the comment space. Relatively little is known about the active promotion of good commenting behaviour, even though the practice by now is widespread among online news platforms. 

As a part of the modern comment section, platforms have been highlighting quality comments on their discussion page \cite{Park2016}. In a practical sense, it takes the form of NYT Picks at the New York Times \cite{Diakopoulos2015b}, Guardian Picks at The Guardian \cite{TheGuardian2009} and featured comments at Dutch news platform NU.nl \cite{NUJij2018}, for instance. Roughly speaking, the platforms define such quality comments as "substantiated", "most interesting and thoughtful" or "presenting a range of perspectives" \cite{NUJij2018,Diakopoulos2015a}. In general, featuring what they deem high-quality content is an attempt by platforms at norm-setting \cite{Wang2022}. Dutch online platform NU.nl specifically states that these comments serve as examples for other users \cite{NUJij2018}.

Even though most research on content moderation focuses on unwanted comments, some have specifically looked at aspects of featured comments. NYT picks in particular have been used as examples of constructive comments in classification tasks \cite{Kolhatkar2017}. Wang and Diakopoulos use a classifier trained on NYT picks to assign quality scores to other comments, concluding that users who receive a NYT pick subsequently write higher quality comments, an effect that diminished over time \cite{Wang2022}. Yahoo News comment threads have been used for the annotation of good content, more specifically in terms of "ERICs: Engaging, Respectful, and/or Informative Conversations" \cite{Napoles2017}. The authors focuses on the thread level rather than on the comment and did not use an editorial standard as their labelling, as is the case with NYT picks \cite{Napoles2017}. Additionally, research has annotated constructive comments as containing specific evidence or solutions, as well as personal anecdotes or stimulating dialogue \cite{Kolhatkar2023}. As part of their visual CommentIQ interface, prior research classifies and ranks comments using comment and user history criteria. These criteria included readability scores and the number of likes a comment had received and relevance scores introduced in earlier research by Diakopoulos \cite{Park2016,Diakopoulos2015a}. The visual interface allows for different plots and ranking possibilities and uses NYT picks \cite{Park2016}. With the similar goal of supporting the moderators to pick featured comments, Waterschoot and Van den Bosch  trained classifiers to rank comments based on the probability that moderators picked them as featured \cite{Waterschoot2024b}. Using data from the Dutch news platform NU.nl, the authors supplemented comment and user information with text representation \cite{Waterschoot2024b}. The models were tested on the discussion level on unseen articles from the platform and evaluated by the NUjij moderators themselves, yielding positive results in regard to the ranking of comments. 

In sum, while literature on online content moderation is mostly aimed at toxic or other unwanted content in the comment space, moderation strategies aimed at promoting good user-generated content are widespread. While previous work did look at practical support for the moderator in picking content and the effect of highlighted comments on the user and replies to the comment, an analysis on the discussion level comparing discussions with featured content to those without has not yet been performed. 
The current study aims to address the potential impact of highlighted quality comments on the discussion by comparing them to a control set of discussions in which this strategy was not performed. Furthermore, we broaden the concept of discussion quality by including a user perspective as well, aside from the editorial definition used in previous work. 

\section{Methodology}
We use a 2023 Dutch language dataset from the comment platform NUjij\footnote{Data available at github.com/cwaterschoot/Featured\_Comments\_Impact}, part of the Dutch online newspaper NU.nl\footnote{https://nu.nl}. The platform allows users to comment on the news articles published by the outlet. 
Aside from the set of articles in which moderators picked featured comments ($n=143$; $86,157$ comments, on average $602$ comments per article, $1,235$ featured comments), we also obtained a control set of articles in which no featured content was chosen ($n=66$; $32,862$ comments, on average $498$ comments per article). Articles in both groups have publication times spread throughout the day. Included in the data are comments that were rejected by the moderators. The articles cover a range of topics such as climate change, the local elections, the nitrogen issue in the Netherlands and the war in Ukraine. Comments are not featured based on the popularity of the news topic. Both the featured group of discussions as well as the control group includes articles covering these topics. In the case of featured comments, a timestamp indicates the exact time that moderators highlighted the comment. 


In order to assess whether the presence of featured content had any impact, each discussion is split in two subgroups: (1) comments before featured content was chosen and (2) comments posted after. For the discussions in which comments were featured by moderators (group Featured), the cut-off was made at the featuring time of the first featured comment per discussion. In the case of control discussions (group Control), the split was made at the median time of these first featured comments relative to the publication time of the article ($123$ minutes). We decided for the median as opposed to the mean ($231$ minutes) due to the impact of articles published late at night, for which the comment sections only opened up in the morning, leading to outliers.

\begin{table}[htb]
\caption{Logistic regression: Characteristics before cut-off}
\small
\centering
\setlength\tabcolsep{2pt}
\begin{tabular}{|l|c|c|}
  \hline
 \textbf{Variable}&  \textbf{Coeff (std er.)} & \textbf{p-value}  \\
 \hline
 Respect count  &   -0.055 (0.039)        & 0.165                      \\
 Featured candidates       &   0.038 (0.030) & 0.209                              \\
 Flagged comments  & 0.034 (0.035) & 0.338                                  \\
 Rejection rate    & -0.006 (0.015) & 0.678                                 \\
 User count  & 0.007 (0.005) & 0.158                                     \\
 Post count     & -0.003 (0.002) & 0.172                                    \\
\hline
\multicolumn{2}{l}{Dependent variable: group (control before/featured before)}
\end{tabular}
\label{reg}
\end{table}

To test the validity of the comparison between the featured and control group, we constructed a logistic regression model based on the before data (Table \ref{reg}). As dependent variable, we included the group identifier, either control or featured. Thus, the model tests whether it is capable to predict if a before discussions belongs to the featured or control group. If this would be possible, the groups show different discussion characteristics. Independent variables are listed in Table \ref{reg}. These variables were averaged across all comments before the cut-off, i.e. for the control group the $123$ minute mark and for the featured group comments posted before the first featuring timestamp in the discussion. Significant effects of these discussion characteristics imply that we cannot conclude that the discussion groups showed similar discussion characteristics before the moderation strategy was performed. This result would suggest the invalidity of our between-group testing. However, the included discussion characteristics were not statistically significant between the two groups (Table \ref{reg}). All comparisons between the after subgroups were made using Mann Whitney U tests with Bonferonni correction to correct for Type I error.

\subsection{Measuring influence: quality \& activity}
We aim to analyze whether discussions in which featured comments are highlighted contain more quality comments compared to discussions without the moderation strategy. We operationalized the concept of \textit{discussion quality} based on two categories, each further broken down into two perspectives. The two categories of discussion quality are (1) the absence of bad content and (2) the presence of quality comments. Each category is analyzed from both the user and editorial (platform) perspective. We contrasted these markers of discussion quality between the before and the after subgroups for the control and featured discussions. 

The absence of bad quality from the user perspective is tested by averaging the percentage of comments that were flagged. A higher rate of flagged comments could be seen as an indication that the user base decided the discussion contained less quality comments. The editorial perspective in this category is operationalized through the rejection rate. This variable captures the percentage of comments in each discussion that moderators decided to delete. Discussion quality would increase if the need to delete unwanted content decreases.

The second category related to discussion quality aims to capture the opposite, i.e. the presence of high quality comments. The user perspective is defined through the average number of likes comments received. For both the before and after subgroups within the control and featured discussions, we calculated the average number of likes. We assessed discussion quality from the editorial perspective by following the procedure outlined in by previous research \cite{Wang2022}. Using a different NUjij dataset containing comments from 2020, we replicated the model for scoring unseen comments in terms of comment quality, operationalized as the class probability for being featured \cite{Wang2022,Waterschoot2024b}. 
The variables used for training the model are described in previous research and include both comment and user features \cite{Waterschoot2024b}. For training and testing, we split the 2020 dataset into an 80/20 split, resulting in $6,679$ featured comments in the training and $1,661$ featured comments in the test set. A random forest model was trained on a balanced set containing the $6,679$ featured comments alongside an equal number of random non-featured comments. The final model, calculated on unseen test set, achieved an F1-score of $0.86$, a similar result as reported in previous work \cite{Wang2022}. Each comment from the current dataset received a probability of being featured-worthy, a proxy for the editorial view of comment quality \cite{Wang2022}. We calculated the averaged percentage of featured-worthy candidates (those with a class probability above $0.5$). 

The final point of focus relates to the evolution of activity within discussions on the platform. We assessed how the discussions evolved in the first $10$ hours by counting the average number of unique users and mean total comments, comparing the before and after subgroups for both the control and featured groups. For this analysis, we only included the accepted comments in the discussion, leaving out the comments that were deleted by a moderator. We are particularly interested in analyzing whether the featured content caused a different evolution in discussion growth.

\section{Results}
Discussion quality was studied comparing the before/after groups and by contrasting the control group with the discussions in which featured content was chosen. We divided quality into two categories: (1) absence of bad content and, (2) presence of high-quality comments. The discussion quality framework resulted in four variables calculated for the before and after subgroups for both the control and feature sets (Table \ref{variables}).

\begin{table*}[htb]
\caption{Discussion Quality: differences between the before and after subgroups}
        \small
        \centering
\begin{tabular}{|l|l|l|c|c|c|c|}
  \hline
\textbf{Category} &  \textbf{Perspective}&\textbf{Variable} & &  \textbf{Before} & \textbf{After} & \textbf{\(\Delta \)} \\
  \hline
Absence of  & User & Flagged comments & Control & 9.08\% & 10.07\% & +0.99pp \\
bad content & & & Feat. & 9.74\% & 9.7\% & -0.04pp \\
  & Editorial& Rejection rate& Control & 23.25\% & 22.20\% & -1.05pp  \\
  & & & Feat. & 23.54\% & 20.91\% & -2.63pp \\
 \hline
  Presence quality& User & Respect Count& Control &  5.69 & 3.61 & -2.08 \\
  comments & & & Feat. & 5.65 & 3.43 & -2.22 \\
   & Editorial & Featured candidates& Control & 14.04\% & 8.90\% & -5.14pp \\
    & & & Feat. & 14.39\% & 8.30\% & -6.09pp \\
\hline
\end{tabular}
\label{variables}
\end{table*}

The user perspective on the absence of bad quality content is expressed through the average percentage of user flagged comments. In the control group, we found that $9.08$\% before the $123$ minute mark and $10.07$\% after the cut-off were flagged by at least one user (+$0.99$pp). For the featured group, we calculated a decrease from $9.74$\% before to $9.70$\% (-$0.04$pp). The results indicate that, while featuring content did not decrease the flagging of bad comments by users, it could prevent more flagged comments later on in the discussion. However, the difference between the average number of flagged comments in the after discussions is not significant ($U=4830.50, p=0.42$). The editorial perspective on absence of bad content was calculated by the average percentage of comments deleted by moderators. Both the control group ($-1.05$pp) and the featured group ($-2.63$pp) showed a decline (Table \ref{variables}). Comparing the after groups of both the featured on control group, we did not find a significant difference ($U=4021.5, p=0.38$). This result suggest that the presence of featured comments did not reduce the need for moderators to reject incoming comments.

The second category of discussion quality aimed to capture the presence of quality content. The user perspective of this category was operationalized by calculating the average number of respect points that comments received before and after the cut-off. Over time, the average number of likes declined, potentially due to the fact that comments posted late in the discussion are read less often. We specifically looked at whether this decline would be less steep in the featured discussion. In the featured group, the average number of likes declined by $2.22$ respect points. In the control group, the average declined by $2.08$ respect points. This difference between the control and featured group after the cut-off is not significant ($U=4420.0, p=0.8$). 

The final marker for discussion quality was the percentage of featured candidates (class probability $>0.5$). The average percentage of candidates before the cut-off showed no difference comparing the featured group ($14.39\%$) and the control group ($14.04$\%). As the discussion continued, the average share of featured candidates comments decreased in both featured ($-6.09pp$) and control ($-5.08pp$) groups (Table \ref{variables}). This difference in average number of featured candidates in the after subgroups were not significant ($U=4093.5, p=0.49$). The average percentage of featured candidates in the featured and control groups after the cut-off were $8.30$ and $8.90$, respectively (Table \ref{variables}). 

Finally, we analyzed discussion activity before and after the cut-off based on two factors: the set of users commenting and, the average number of comments in the discussions. This analysis omitted those that were rejected by moderators. 

\begin{figure*}
\centering
\begin{subfigure}{.5\columnwidth}
    \centering
    \includegraphics[width=1\columnwidth]{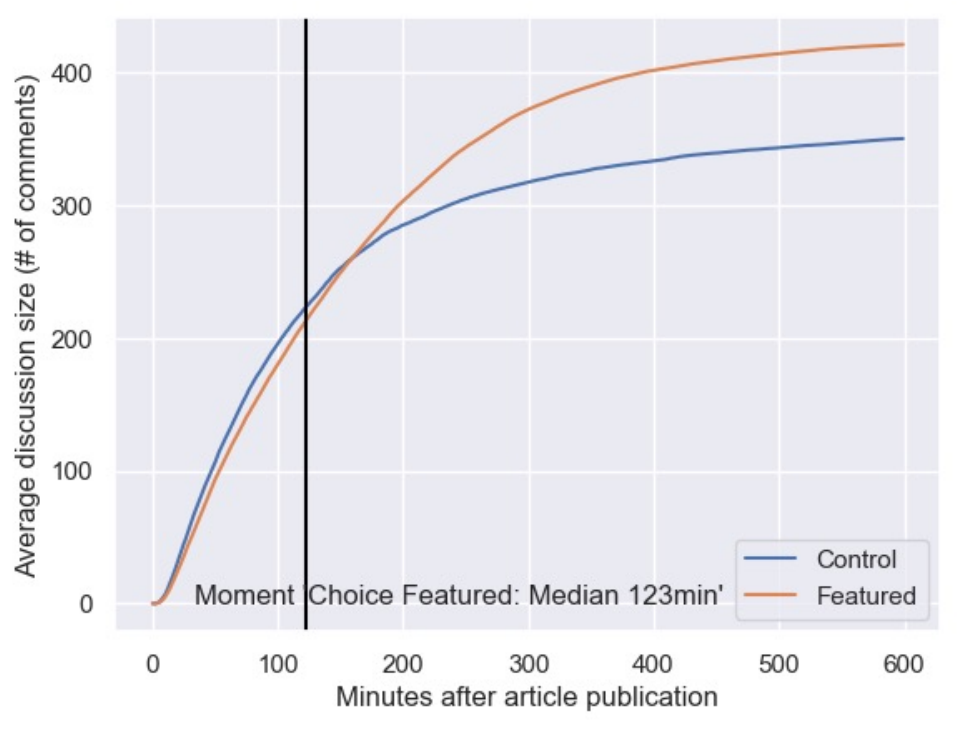}
    \caption{Evolution of average post count}
    \label{postcount}
\end{subfigure}%
\begin{subfigure}{.5\columnwidth}
    \centering
    \includegraphics[width=1\columnwidth]{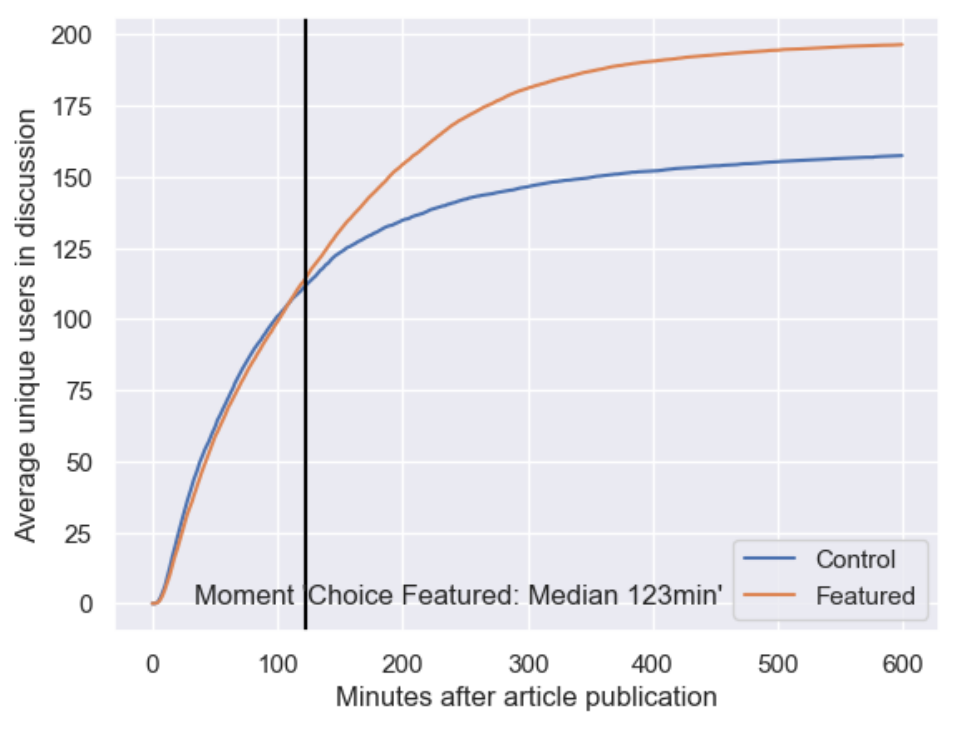}
    \caption{Evolution of unique users}
    \label{users}
\end{subfigure}
\caption{Growth of discussion activity in featured and control group (first 600min)}
\label{activity-full}
\end{figure*}

Unlike the quality markers discussed earlier, the discussion activity progressed differently within the featured group as opposed to the control discussions. The speed of discussion activity slowed down over time, which is expected due to the fact that users move on to more recent articles. However, the average discussion activity in the control group dwindled down much quicker; the featured discussions continued on and slowed down at a point later on (Fig. \ref{activity-full}). 

Before the cut-off, an average of $112$ users commented before the $123$ minute mark in the control discussions, while $115$ users participated before the first comment was featured in the featured group. In the case of the after groups, however, a significant difference was found, indicating that more unique users commented in the featured discussions after content was featured ($U=3205.5, p<0.001$). The same result was found for the post count. The average post count before the $123$ minute mark in the control discussions was $223$, while in the featured group this was $213$ comments. After the cut-off, the average control discussion went on for $126$ more comments, while the mean post count in the featured group was $207$. This difference in after groups is statistically significant ($U=3161.5, p<0.001$).

\section{Discussion}

In the following sections, we discuss the apparent inability to influence discussion quality with featured comments. We end the discussion by outlining several open questions for future work and the limitations of the current study. 


First, featuring content did not lower the necessity to reject incoming comments, implying that the highlighted examples did not deter people from posting uncivil or off-topic content. While the rejection rate did decrease over time, control discussions without featured comments evolved in the same manner. From the user perspective, the control group experienced an increase in user flagging in the after group, while this increase was not found in the group with featured comments. However, this difference was not significant. 

Second, we found no difference between the two groups in regard to the user perspective on quality, represented in the study by the number of likes comments received. Over time, the average number of likes a comment received decreased significantly with or without featured comments. The final marker of discussion quality entailed featured-worthy comments. If such contributions would successfully serve as examples to users of the editorial standard of quality, one would expect the number of comments qualifying as featured-worthy to be significantly higher in the after group of featured discussion compared to the after group comprising of control discussions. However, we did not find such a significant difference. In both groups, the number of featured-worthy comments decreased in the after subgroup. Overall, these results indicate that discussion quality decreased over time. 
Even though the practice of featuring user comments is widespread among online news platforms, the inability to influence discussion quality in all but one of the variables (and not significantly so) indicates that users do not use these comments as examples. In particular the lack of change in the editorial perspective showcased that the platform does not succeed in shaping the discussion to what they themselves deem good discussion. 



We hypothesize that featured content could be used to postpone the natural decrease in discussion activity. Further studies should focus on this particular point, eliminating other factors through, among other things, A/B testing. Additionally, cross-platform analysis is necessary to assess the general impact of the moderation strategy. Platforms like, for example, the New York Times and the Guardian highlight certain comments as well. Such an analysis can compare platforms to test whether moderators at different outlets feature similar content.

Research aimed at analyzing the effects of online moderation has the inherent constraint of the availability of data and metadata. To replicate the current study, researchers require not only the comments that were published at the time, but also information about those which were rejected or later deleted by moderators. The latter is not publicly available, typically, cannot be published, and requires cooperation with online news platforms to be obtained, or working from within the platform's organization. Furthermore, individual comments require metadata indicating to which article they were posted, such that researchers can separate different discussions. Timestamps indicating when comments were featured are also not publicly available, typically, as well as information as to how many times a comment might have been flagged by other users. All in all, without this crucial information, the potential effect of these moderation strategies cannot be adequately analyzed. Cooperation with the platforms is needed to obtain such unpublished information.

\section{Conclusion}
To sum up, we did not find evidence indicating an impact on discussion quality as a result of the featured content, especially from the editorial perspective. The rejection percentage was unaffected, as well as the number of featured-worthy comments. Both aspects of discussion quality decreased in both the control and featured groups in similar fashion. A similar decrease was found in the average respect points comments had received, the user perspective of presence of quality comments. What did change, however, was the average number of flagged comments by users. As opposed to the other quality variables in the framework, the average number of flagged comments increased in the after group of the control discussion, while this increase was not found in the featured group. However, this difference was not statistically significant between after groups.

Finally, we did find differences in discussion activity between the featured and control group. The results show that discussion activity declined slower in the featured group. It would seem that this moderation strategy can be used to postpone the decline in user activity. However, future research is necessary to eliminate other factors potentially influencing discussion growth.

\begin{credits}
\subsubsection{\ackname} This study is financed by project number 410.19.006 of the research program `Digital Society - The Informed Citizen' which is financed by the Dutch Research Council (NWO).

\subsubsection{\discintname}
The authors have no competing interests to declare that are
relevant to the content of this article.
\end{credits}
%
%
%
\bibliographystyle{splncs04}
\bibliography{library}

\begin{thebibliography}{10}
\providecommand{\url}[1]{\texttt{#1}}
\providecommand{\urlprefix}{URL }
\providecommand{\doi}[1]{https://doi.org/#1}

\bibitem{Diakopoulos2015b}
Diakopoulos, N.: {Picking the NYT Picks : Editorial Criteria and Automation in the Curation of Online News Comments}. {\#}ISOJ, the official research journal of ISOJ  \textbf{5}(1),  147--166 (2015)

\bibitem{Diakopoulos2015a}
Diakopoulos, N.: {The editor's eye: Curation and comment relevance on the New York Times}. CSCW 2015 - Proceedings of the 2015 ACM International Conference on Computer-Supported Cooperative Work and Social Computing pp. 1153--1157 (2015). \doi{10.1145/2675133.2675160}

\bibitem{Gillespie2018}
Gillespie, T.: {Custodians of the Internet: Platforms, content moderation, and the hidden decisions that shape social media}. Yale University Press (2018)

\bibitem{Gillespie2022}
Gillespie, T.: {Do Not Recommend? Reduction as a Form of Content Moderation}. Social Media and Society  \textbf{8}(3),  1--13 (2022). \doi{10.1177/20563051221117552}

\bibitem{Kolhatkar2017}
Kolhatkar, V., Taboada, M.: {Constructive language in news comments}. Proceedings of the Annual Meeting of the Association for Computational Linguistics pp. 11--17 (2017). \doi{10.18653/v1/w17-3002}

\bibitem{Kolhatkar2023}
Kolhatkar, V., Thain, N., Sorensen, J., Dixon, L., Taboada, M.: {Classifying constructive comments}. First Monday  \textbf{28}(4),  1--16 (2023). \doi{https://doi.org/10.5210/fm.v28i4.13163}

\bibitem{Lewandowsky2017}
Lewandowsky, S., Ecker, U.K., Cook, J.: {Beyond Misinformation: Understanding and Coping with the “Post-Truth” Era}. Journal of Applied Research in Memory and Cognition  \textbf{6}(4),  353--369 (2017). \doi{10.1016/j.jarmac.2017.07.008}

\bibitem{vanderlinden2024}
van~der Linden, L., Waterschoot, C., van~den Hemel, E., Kunneman, F., Bosch, A.v.d., Krahmer, E.: Who are the online commenters: A large-scale representative survey to explore the identity and motivation of online commenters (Jun 2024). \doi{10.31234/osf.io/fxjkw}, \url{osf.io/preprints/psyarxiv/fxjkw}

\bibitem{VanderLinden2017}
van~der Linden, S., Leiserowitz, A., Rosenthal, S., Maibach, E.: {Inoculating the Public against Misinformation about Climate Change}. Global Challenges  \textbf{1}(2),  1600008 (2017). \doi{10.1002/gch2.201600008}

\bibitem{Napoles2017}
Napoles, C., Tetreault, J., Rosato, E., Provenzale, B., Pappu, A.: {Finding good conversations online: The yahoo news annotated comments corpus}. LAW 2017 - 11th Linguistic Annotation Workshop, Proceedings of the Workshop pp. 13--23 (2017). \doi{10.18653/v1/w17-0802}

\bibitem{NUJij2018}
NUJij: {NUjij - Veelgestelde vragen} (2018), \url{https://www.nu.nl/nujij/5215910/nujij-veelgestelde-vragen.html}

\bibitem{Paasch2022}
Paasch-Colberg, S., Strippel, C.: {“The Boundaries are Blurry{\ldots}”: How Comment Moderators in Germany See and Respond to Hate Comments}. Journalism Studies  \textbf{23}(2),  224--244 (2022). \doi{10.1080/1461670X.2021.2017793}

\bibitem{Park2016}
Park, D., Sachar, S., Diakopoulos, N., Elmqvist, N.: {Supporting comment moderators in identifying high quality online news comments}. Conference on Human Factors in Computing Systems - Proceedings pp. 1114--1125 (2016). \doi{10.1145/2858036.2858389}

\bibitem{Quandt2018}
Quandt, T.: {Dark participation}. Media and Communication  \textbf{6}(4),  36--48 (2018). \doi{10.17645/mac.v6i4.1519}

\bibitem{Roberts2017}
Roberts, S.T.: {Content Moderation}. Encyclopedia of Big Data pp.~1--4 (2017). \doi{10.1201/9781003293125-7}

\bibitem{TheGuardian2009}
{The Guardian}: {Frequently asked questions about community on the Guardian website} (2009), \url{https://www.theguardian.com/community-faqs}

\bibitem{Wang2022}
Wang, Y., Diakopoulos, N.: {Highlighting High-quality Content as a Moderation Strategy : The Role of New York Times Picks in Comment Quality}. ACM Transactions on Social Computing  \textbf{4}(4),  1--24 (2022)

\bibitem{Waterschoot2024}
Waterschoot, C.: {Governing the 'Third Half of the Internet': The Dynamics of Human and AI-assisted Content Moderation}. In: van Dijck, J., van Es, K., Helmond, A., van~der Vlist, F. (eds.) Governing the Digital Society: Platforms, Artificial Intelligence, and Public Values. Amsterdam University Press (2024), forthcoming

\bibitem{Waterschoot2024b}
Waterschoot, C., van~den Bosch, A.: {A time-robust group recommender for featured comments on news platforms}. Frontiers in Big Data  \textbf{7} (2024). \doi{10.3389/fdata.2024.1399739}

\bibitem{Wintterlin2020}
Wintterlin, F., Schatto-Eckrodt, T., Frischlich, L., Boberg, S., Quandt, T.: {How to Cope with Dark Participation: Moderation Practices in German Newsrooms}. Digital Journalism  \textbf{8}(7),  904--924 (2020). \doi{10.1080/21670811.2020.1797519}

\bibitem{Wolfgang2018}
Wolfgang, J.D.: {Cleaning up the “Fetid Swamp”: Examining how journalists construct policies and practices for moderating comments}. Digital Journalism  \textbf{6}(1),  21--40 (2018). \doi{10.1080/21670811.2017.1343090}

\end{thebibliography}

\end{document}